\pdfoutput=1

\documentclass[11pt]{article}

\usepackage[hyperref]{EMNLP2022}

\usepackage{times}
\usepackage{latexsym}

\usepackage{times}
\usepackage{latexsym}

\usepackage{enumitem}
\usepackage{multirow}
\usepackage{array}
\usepackage{booktabs} 
\usepackage{microtype}
\usepackage{times}
\usepackage{latexsym}
\usepackage{multirow}
\usepackage{graphics}
\usepackage{booktabs} 
\usepackage{bm}
\usepackage{url}
\usepackage{mathrsfs}
\usepackage{multirow}
\usepackage{balance}
\usepackage{amsthm}
\usepackage{amsmath}
\usepackage{amstext}
\usepackage{amssymb}
\usepackage[ruled,vlined,linesnumbered]{algorithm2e}
\usepackage{subcaption}
\usepackage{overpic}
\usepackage{color}
\usepackage{enumitem}
\usepackage{array}
\usepackage{booktabs} 
\usepackage{enumitem}

\usepackage{microtype}
\usepackage[T1]{fontenc}
\usepackage[utf8]{inputenc}

\title{Robustness Challenges in Model Distillation and Pruning for Natural Language Understanding}

\author{Mengnan Du\textsuperscript{1}\thanks{\, Most of the work was completed while the first author was an intern at Microsoft Research during summer 2021.}, Subhabrata Mukherjee\textsuperscript{2}, Yu Cheng\textsuperscript{2}, Milad Shokouhi\textsuperscript{2}, \\  \textbf{Xia Hu\textsuperscript{3}, Ahmed Hassan Awadallah\textsuperscript{2}}\\
  \textsuperscript{1}New Jersey Institute of Technology \,
  \textsuperscript{2}Microsoft Research \,
  \textsuperscript{3}Rice University\\
  \small\texttt{mengnan.du@njit.edu}, \small\texttt{xia.hu@rice.edu} \\
  \small\texttt{\{submukhe,Yu.Cheng,milads,hassanam\}@microsoft.com}
}

\date{}

\begin{document}
\maketitle
\begin{abstract} 
Recent work has focused on compressing pre-trained language models (PLMs) like BERT where the major focus has been to improve the in-distribution performance for downstream tasks. However, very few of these studies have analyzed the impact of compression on the generalizability and robustness of compressed models for out-of-distribution (OOD) data. Towards this end, we study two popular model compression techniques including knowledge distillation and pruning and show that the compressed models are significantly less robust than their PLM counterparts on OOD test sets although they obtain similar performance on in-distribution development sets for a task. Further analysis indicates that the compressed models overfit on the shortcut samples and generalize poorly on the hard ones. We further leverage this observation to develop a regularization strategy for robust model compression based on sample uncertainty. 
Experimental results on several natural language understanding tasks demonstrate that our bias mitigation framework improves the OOD generalization of the compressed models, while not sacrificing the in-distribution task performance.
\end{abstract}

\section{Introduction}
Large pretrained language models (PLMs) (e.g., BERT~\cite{devlin2018bert}, RoBERTa~\cite{liu2019roberta}, GPT-3~\cite{brown2020language}) have obtained state-of-the-art performance in several Natural Language Understanding (NLU) tasks. However, recent studies~\cite{niven2019probing,du2021towards,mudrakarta2018did} indicate that PLMs heavily rely on \underline{shortcut learning/spurious correlations}, rather than acquiring higher level language understanding and semantic reasoning in several NLU tasks. Specifically, these models often exploit dataset biases and artifacts, e.g., lexical bias and overlap bias, as shortcuts for prediction. Due to the independent and identically distributed (IID) split of training, development, and test sets, these models that learn spurious decision rules from training data can perform well on in-distribution data~\cite{du2022shortcut}. Nevertheless, the shortcut learning behavior will result in models that have poor generalization performance on out-of-distribution (OOD) data, raising concerns about their robustness.

On the other hand, it is difficult to use these large PLMs models in real-world applications with latency and capacity constraints, e.g., on edge devices and mobile phones. Thus, model compression emerges as one of the techniques to reduce model size, speed up inference, and save energy without significant performance drop for downstream tasks. State-of-the-art model compression techniques such as knowledge distillation~\cite{sanh2019distilbert,sun2019patient} and pruning~\cite{sanh2020movement} primarily focus on evaluating compressed model performance in in-distribution test data. However, in-distribution testing is insufficient to capture the generalizability of PLMs~\cite{d2020underspecification}. In contrast to existing work that is geared towards general-purpose PLMs~\cite{niven2019probing,du2021towards,mudrakarta2018did}, this work aims to study the impact of compression on the shortcut learning and OOD generalization ability of compressed models. 

Towards this end, we conduct comprehensive experiments to evaluate the OOD robustness of compressed models, with BERT as the base encoder. \textbf{We focus primarily on two popular model compression techniques in the form of pruning and knowledge distillation}~\cite{sanh2019distilbert,wang2020minilm}. For pruning, we consider two popular techniques including iterative magnitude pruning~\cite{sanh2020movement} and structured pruning~\cite{prasanna2020bert,liang2021super}. Specifically, we explore the following research questions: \emph{Are distilled and pruned models as robust as their PLM counterparts for downstream NLU tasks? What is the impact of varying the level of compression on OOD generalization and bias of compressed models?} We evaluate the performance of several compressed models obtained using the above techniques on both standard in-distribution development sets and OOD test sets for downstream NLU tasks. Experimental analysis indicates that distilled and pruned models are consistently less robust than their PLM counterparts. Further analysis of the poor generalization performance of compressed models reveals some interesting observations. For instance, we observe that the compressed models overfit on the easy / shortcut samples and generalize poorly on the hard ones. This motivates our second research question: \emph{How to regularize model compression techniques to generalize across samples with varying difficulty?} This brings some interesting challenges since we do not know which samples are easy or hard apriori. 

Based on the above observations, we propose a bias mitigation framework to improve the OOD robustness of compressed models, termed as RMC (\underline{R}obust \underline{M}odel \underline{C}ompression). First, we leverage the uncertainty of the deep neural network to quantify the difficulty of a training sample. This is given by the variance in the prediction of a sample from multiple sub-networks of the original large network obtained by model pruning. Second, we leverage this sample-specific measure for smoothing and regularizing different families of compression techniques. 
The major contributions of this work can be summarized as follows:
\begin{itemize}[leftmargin=*]\setlength\itemsep{-0.3em}
\item We perform a comprehensive analysis to evaluate the OOD generalization ability and robustness of compressed models for NLU tasks. 
\item We further analyze plausible reasons for the low generalizability of compressed models and demonstrate connections to shortcut learning.
\item We propose a mitigation framework for regularizing model compression, termed as RMC, which   
smooths the knowledge distillation training based on the estimated sample difficulties.

\item We perform experiments to demonstrate that our RMC framework improves OOD generalization while not sacrificing the standard in-distribution task performance on multiple NLU tasks.
\end{itemize}

\section{Related Work}

\noindent\textbf{Shortcut Learning and Mitigation.}\, 
Recent studies indicate that PLMs tend to exploit biases and artifacts in the dataset as shortcuts for prediction, rather than acquiring higher level semantic understanding and reasoning for NLU tasks~\cite{niven2019probing,du2021towards,mccoy2019berts}. There are some preliminary work to mitigate the bias of general PLMs, including product-of-experts~\cite{clark2019don,he2019unlearn,sanh2020learning}, re-weighting~\cite{schuster2019towards,yaghoobzadeh2019robust,utama2020towards}, adversarial training~\cite{stacey2020there}, posterior regularization~\cite{cheng2020posterior}, etc. 

\vspace{2pt}
\noindent\textbf{Robustness in Model Compression.}\,  
Current practice for evaluating model compression performance focuses mainly on standard benchmark performance \cite{Zhu2020FreeLB,wang2021infobert}. In the computer vision domain, previous work shows that compressed models perform poorly in Compression Identified Exemplars (CIE)~\cite{hooker2019compressed}, and compression amplifies algorithmic bias towards certain demographics~\cite{hooker2020characterising}. 
The most similar work to ours are two concurrent work~\cite{xu2021beyond,li2021select} that investigate the performance of compressed models beyond standard benchmarks for natural language understanding tasks. However, both work mainly focus on evaluating the robustness of compressed models with respect to the scenario of adversarial attacks, i.e., TextFooler~\cite{jin2019bert}, and the unified adversarial framework~\cite{li2021select}. In contrast, we comprehensively characterize the robustness of BERT compression in OOD test sets to probe the OOD generalizability of the compression techniques. Besides, we use insights from this robustness analysis to design a generalizable and robust model compression framework.

\section{Are Compressed Models Robust?}
We perform a comprehensive analysis to evaluate the robustness of compressed language models.

\subsection{Compression Techniques}\label{sec:three-families-of-compression}
We consider two popular families of compression, namely, knowledge distillation and model pruning.

\vspace{2pt}
\noindent\textbf{Knowledge Distillation}: The objective here is to train a small-size model by mimicking the behavior of the larger teacher model using knowledge distillation~\cite{hinton2015distilling}. In this work, we focus on task-agnostic distillation. In particular, we consider DistilBERT~\cite{sanh2019distilbert} and MiniLM~\cite{wang2020minilm} distilled from BERT-base. For a fair comparison, we select compressed models with similar capacities ($66M$ parameters in this work). In order to evaluate the impact of compression techniques on model robustness, we also consider similar capacity smaller models without using knowledge distillation. These are obtained via simple truncation where we retain the first $6$ layers of the large model, and via pre-training a smaller $6$-layer model from scratch. 

\vspace{2pt}
\noindent\textbf{Iterative Magnitude Pruning}: This is a task-specific unstructured pruning method~\cite{sanh2020movement}. During the fine-tuning process for each downstream task, the weights with the lowest magnitude are removed until the pruned model reaches the target sparsity. Note that we utilize the standard pruning technique, rather than the LTH-based pruning (lottery ticket hypothesis) that uses re-winding~\cite{chen2020lottery}. We also consider different pruning ratios to obtain pruned models with different levels of sparsity. 

\vspace{2pt}
\noindent\textbf{Structured Pruning}: This method family is based on the hypothesis that there is redundancy in the attention heads~\cite{prasanna2020bert,voita2019analyzing,bian2021attention,chen2021chasing}. We also consider task-specific pruning. During the fine-tuning process for each task, it prunes the whole attention heads based on their importance to the model predictions. Please refer to Sec.~\ref{Details-of-Pruning-Methods} in Appendix for more details.
We prune around $20\%$ attention heads in total (i.e., $28$ attention heads). Further pruning increases the sparsity with significant degradation of the model's performance on in-distribution development sets.

\subsection{Evaluation Datasets}\label{Evaluation-Datasets}
To evaluate the robustness of the compressed models introduced in the last section, we use three NLU tasks, including MNLI, FEVER, and QQP\footnote{MNLI, FEVER, and QQP are the three most widely used datasets to evaluate the shortcut learning/bias behavior and OOD generalization of PLMs in the literature~\cite{tu2020empirical,he2019unlearn,clark2019don,schuster2019towards}}. Please refer to Sec.~\ref{More-on-Evaluation-Datasets} in Appendix for more details.
\begin{itemize}[leftmargin=*]\setlength\itemsep{-0.3em}
\item \noindent\textbf{MNLI}~\cite{williams2017broad}: This is a natural language inference task. In this work, we report the accuracy metric on the matched subset. We use HANS~\cite{mccoy2019right} as the adversarial test set, which contains $30,000$ synthetic samples. Models that exploit shortcut features have been shown to perform poorly on the HANS test set.

\begin{table*}
\centering
\setlength\tabcolsep{3.5pt}
\scalebox{0.72}{
\begin{tabular}{l c c c c c c c c c c c c |c}
\toprule 
& & \multicolumn{3}{c}{\textbf{MNLI}} & \multicolumn{4}{c}{\textbf{FEVER}} & \multicolumn{4}{c|}{\textbf{QQP}} \\
\cmidrule(l){3-5} \cmidrule(l){6-9} \cmidrule(l){10-13}
\textbf{Sparsity} &\#Param & DEV & HANS &$\mathcal{F}_{bias}$  & DEV & Sym1 & Sym2  &$\mathcal{F}_{bias}$ & DEV & $\text{paws}_{wiki}$ & $\text{paws}_{qqp}$  &$\mathcal{F}_{bias}$ & Average $\mathcal{F}_{bias}$\\ 
\midrule 
BERT-base &109M & 84.2 & 59.8 &-      & 86.2 & 58.9 & 64.5 &-     & 90.9 & 48.9 & 34.7&-    &-\\\midrule 
20\%      &87.2M & 84.4 & 55.5 &1.182  & 86.5 & 57.0 & 64.6 &1.045 & 90.7 & 47.2 & 33.5&1.037&1.088\\
40\%      &65.4M & 84.0  & 54.7 &1.204 & 86.4 & 57.2 & 64.0 &1.051 & 90.5 & 46.6 & 32.4&1.049&1.101\\
60\%      &43.6M & 83.4 & 52.8  &1.266 & 86.3 & 56.9 & 63.3 &1.068 & 90.2 & 45.9 & 31.8&1.061&1.132\\
70\%      &32.7M & 81.8 & 52.2  &1.249 & 85.9 & 56.6 & 63.3 &1.063 & 89.5 & 45.4 & 30.7&1.065&1.127\\
\bottomrule 
\end{tabular}
}
\vspace{-4pt}
\caption{Accuracy comparison (in percent) and relative bias $\mathcal{F}_{bias}$ (the smaller the better) for models with iterative magnitude pruning with different levels of sparsity. The last column indicates the average $\mathcal{F}_{bias}$ values over three tasks. \emph{Pruned models have relatively higher degradation in OOD test set compared to the development set.} 
}
\label{tab:iterative-magnitude-pruning}
\end{table*}

\begin{table*}
\centering
\setlength\tabcolsep{3.5pt}
\scalebox{0.72}{
\begin{tabular}{l c c c c c c c c c c c c| c}
\toprule 
& & \multicolumn{3}{c}{\textbf{MNLI}} & \multicolumn{4}{c}{\textbf{FEVER}} & \multicolumn{4}{c|}{\textbf{QQP}} \\
\cmidrule(l){3-5} \cmidrule(l){6-9} \cmidrule(l){10-13}
\textbf{Sparsity} &\#Param & DEV & HANS &$\mathcal{F}_{bias}$  & DEV & Sym1 & Sym2  &$\mathcal{F}_{bias}$ & DEV & $\text{paws}_{wiki}$ & $\text{paws}_{qqp}$  &$\mathcal{F}_{bias}$ & Average $\mathcal{F}_{bias}$\\
\midrule 
BERT-base  &109M& 84.2 & 59.8 &-      & 86.2 & 58.9 & 64.5 &-     & 90.9 & 48.9 & 34.7&-&-\\\midrule
DistilBERT &66M & 82.3 & 51.2 &1.289  & 84.5 & 51.9 & 60.4 &1.183 & 89.9 & 48.1 & 34.6& 1.006 &1.159\\
MiniLM     &66M & 83.1 & 51.4 &1.309  & 84.2 & 53.4 & 60.7 &1.137 & 89.9 & 46.8 & 31.0& 1.039  &1.162\\
Truncated-l6 &66M &80.8 & 51.6&1.247  & 84.4 & 52.6 & 60.4 &1.163 & 90.0 & 46.0 & 32.4& 1.056  &1.155\\
Pretrained-l6 &66M &81.6 &52.2&1.229  & 85.8 & 54.7 & 62.6 &1.115 & 90.0 & 46.4 & 33.9& 1.045  &1.130\\
\bottomrule 
\end{tabular}}
\vspace{-4pt}
\caption{Accuracy comparison (in percent) and relative bias $\mathcal{F}_{bias}$ (the smaller the better) of compressed models with knowledge distillation. \emph{Distilled models have relatively higher degradation in OOD test set compared to the development set.} Except BERT-base, all other models have $66M$ parameters. }
\label{tab:knowledge-distillation}
\end{table*}

\item \noindent\textbf{FEVER}~\cite{thorne2018fever}: This is a fact verification dataset. Recent studies indicate that there are strong shortcuts in the claims~\cite{utama2020towards}. To facilitate the robustness and generalization evaluation of fact verification models, two symmetric test sets (i.e., Sym v1 and Sym v2) were created, where bias exists in the symmetric pairs~\cite{schuster2019towards}. Both OOD test sets have $712$ samples.

\item \noindent\textbf{QQP}: The task is to predict whether a pair of questions is semantically equivalent. We consider the OOD test set PAWS-qqp, which contains $677$ test samples generated from QQP corpus~\cite{zhang2019paws,pawsx2019emnlp}. Besides, we also consider the PAWS-wiki OOD test set, which consists of $8,000$ test samples generated from Wikipedia pages.

\end{itemize}

For all three tasks, we employ accuracy as the evaluation metric and evaluate the performance of the compressed models on both the in-distribution development set and the OOD test set. 

\subsection{Evaluation Setup}\label{sec:Evaluation-Analysis-robustness}
In this work, we use the uncased BERT-base as the teacher network, and study the robustness of its compressed variants. The final model consists of the BERT-base encoder (or its compressed variants) with a classification head (a linear layer on top of the pooled output). 
Recent studies indicate that factors such as learning rate and training epochs could have a substantial influence on robustness~\cite{tu2020empirical}. In particular, increasing training epochs can help improve the generalization of the OOD test set. In this work, we focus on the \underline{relative robustness} of compressed models compared to the uncompressed teacher, rather than their \emph{absolute accuracies}. For a fair comparison, we unify the experimental setup for all models. We use Adam optimizer with weight decay~\cite{loshchilov2017decoupled}, where the learning rate is fixed as $2e$-5, and we train all models for 5 epochs on all datasets. We perform the experiments using PyTorch and use the pre-trained models from the Huggingface model pool~\cite{wolf2019huggingface}.
We report the average results over three runs for all experiments.

\begin{table}
\centering
\scalebox{0.7}{
\begin{tabular}{l c c c c}
\toprule 
\textbf{Models} & Attemtion heads & DEV & HANS &$\mathcal{F}_{bias}$ \\ 
\midrule 
BERT-base        & 144 & 84.2   & 59.8  &-\\ \midrule
BERT-116heads-v1 & 116 & 84.1   & 55.5 & 1.172\\
BERT-116heads-v2 & 116 & 84.2   & 53.7 & 1.250 \\
BERT-116heads-v3 & 116 & 84.0   & 55.3  & 1.176\\
\bottomrule 
\end{tabular}}
\vspace{-4pt}
\caption{Accuracy comparison (in percent) and relative bias $\mathcal{F}_{bias}$ (the smaller the better) of compressed models with structured pruning. \emph{Pruned models have relatively higher degradation in OOD test set compared to the development set.} All compressed models have been pruned 28 attention heads. }
\label{tab:structured-pruning}
\end{table}

\subsection{Relative Robustness Metric}
As we later demonstrate, with increase in compression ratio or model sparsity, the performance of the smaller models degrades for both in-domain and OOD test sets. To compare the gap between in-distribution task performance and OOD generalizability, we define a new metric that measures this performance gap of the compressed models with respect to the uncompressed BERT-base (teacher). First, we calculate the accuracy gap between in-distribution development set and OOD test set as $\frac{F_{\text{dev}}-F_{\text{OOD}}}{F_{\text{dev}}}$ for BERT-base (denoted by $\Delta_{\text{BERT-base}}$); and its compressed variant (denoted by $\Delta_{\text{compressed}}$). Second, we compute the relative bias as the ratio between the accuracy gap of the compressed model with respect to BERT-base: 
$
\mathcal{F}_{bias} = \frac{\Delta_{\text{compressed}}}{\Delta_{\text{BERT-base}}}.
$
Here $\mathcal{F}_{bias} > 1$ indicates that the compressed model is more biased than BERT-base with the degree of bias captured in a larger value of $\mathcal{F}_{bias}$.
Since FEVER has two OOD test sets, we use the overall accuracy of sym1 and sym2 to calculate $\mathcal{F}_{bias}$. Similarly, the OOD accuracy for QQP is the overall accuracy on PAWS-wiki and PAWS-qqp.

\subsection{Experimental Observations} 
We report the performance of accuracy and the relative bias measure $\mathcal{F}_{bias}$ for iterative magnitude pruning in Table~\ref{tab:iterative-magnitude-pruning}, knowledge distillation in Table~\ref{tab:knowledge-distillation} and structured pruning in Table~\ref{tab:structured-pruning}. 
We have the following key observations.

\vspace{2pt}
\noindent \textbf{Iterative Magnitude Pruning}: First, for slight and mid-level sparsity, the pruned models have comparable and sometimes even better performance on the in-distribution development set. Consider FEVER as an example, where the compressed model preserves the accuracy on the in-distribution set even at 60\% sparsity\footnote{Here, 60\% sparsity indicates that 40\% parameters are remaining after pruning.}. However, the generalization accuracy on the OOD test set has a substantial drop. This indicates that the development set fails to capture the generalizability of the pruned models. Second, as the sparsity increases, the generalization accuracy on the OOD test set substantially decreases while dropping to random guess for tasks such as MNLI. Third, at high levels of sparsity (e.g. $70\%$), both development and OOD test set performances are significantly affected. In general, we observe $\mathcal{F}_{bias}>1$ for all levels of sparsity in Table~\ref{tab:iterative-magnitude-pruning}. Note that we limit the maximum sparsity at $70\%$ after which the training is unstable with a significant performance drop even on the development set~\cite{liang2021super}. As in the previous cases, there is substantial accuracy drop on the OOD test set compared to the development set (e.g., $7.6\%$ vs $1.9\%$ degradation respectively for the MNLI task).   

\vspace{2pt}
\noindent \textbf{Knowledge Distillation}: Similar to pruning, we observe a higher accuracy drop in the OOD test set compared to the in-distribution development set for distilled models. Consider DistilBERT performance on MNLI as an example with $1.9\%$ accuracy drop in development set compared to $8.6\%$ drop in the OOD test set. This can also be validated in Table~\ref{tab:knowledge-distillation}, where all $\mathcal{F}_{bias}$ values are larger than $1$, depicting that all the distilled models are less robust than BERT-base. Another interesting observation is that distilled models, i.e., DistilBERT and MiniLM, have higher bias $\mathcal{F}_{bias}$ compared to the pre-trained models, i.e., Pretrained-l6 and Truncated-l6, as we compare their average $\mathcal{F}_{bias}$ values in Table~\ref{tab:knowledge-distillation}. This indicates that the compression process plays a significant role in the low generalizability and robustness of the distilled models.

\vspace{2pt}
\noindent \textbf{Structured Pruning}: Recent studies have reported the super ticket phenomenon~\cite{liang2021super}. The authors observe that, when the BERT-base model is slightly pruned, the accuracy of the pruned models improves on in-distribution development set. However, we observe that this finding does not hold for OOD test sets. From Table~\ref{tab:structured-pruning}, we observe that all pruned models are less robust than BERT-base, with $\mathcal{F}_{bias}$ much larger than $1$.

\section{Attribution of Low Robustness}\label{sec:Attribution-of-Low-Robustness}
In this section, we explore the factors that lead to low robustness of compressed models. Previous work has demonstrated that the performance of different models on the GLUE benchmark~\cite{wang-etal-2018-glue} tends to correlate with the performance on MNLI, making it a good representative of natural language understanding tasks in general~\cite{phang2018sentence,liu2020mmtdnn}.
For this reason, we choose the MNLI task for a study.

For the MNLI task, we consider the dataset splits from ~\cite{gururangan2018annotation}. The authors partition the development set into easy/shortcut~\footnote{We use `easy' and `shortcut' interchangeably in this work.} and hard subsets. 
In this experiment, we use pruned models with varying sparsity to investigate the reason for the low robustness of the compressed models. We have the following key observations.

\begin{figure}
  \centering
  \includegraphics[width=0.98\linewidth]{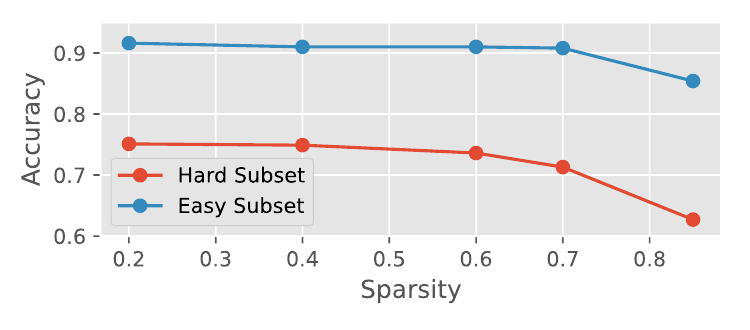}
  \vspace{-3mm}
  \caption{Pruned model performance on hard vs easy / shortcut samples with varying sparsity, where x-axis denotes the sparsity level.}
  \label{fig:attribution}
\end{figure}

\vspace{3pt}
\noindent\textbf{Observation 1}: The compressed models tend to overfit the easy/shortcut samples and generalize poorly on the hard ones. The performance of pruned models at five sparsity levels (ranging between $[0.2 - 0.85]$) on the easy and hard samples for the MNLI task is illustrated in Figure~\ref{fig:attribution}. It demonstrates that the accuracy on the hard samples is much lower compared to the accuracy on the easy ones. As the sparsity increases, we observe a larger accuracy drop on the hard samples compared to the easy ones. In particular, the accuracy gap between the two subsets is $22.7\%$ at the sparsity of $0.85$, much higher than the $16.1\%$ accuracy gap at the sparsity of $0.4$. 
These findings demonstrate that the compressed models overfit on the easy samples, while generalizing poorly on the hard ones. Furthermore, this phenomenon is amplified at higher levels of sparsity for the pruned models. 

\begin{figure*}
  \centering
  \includegraphics[width=0.95\linewidth]{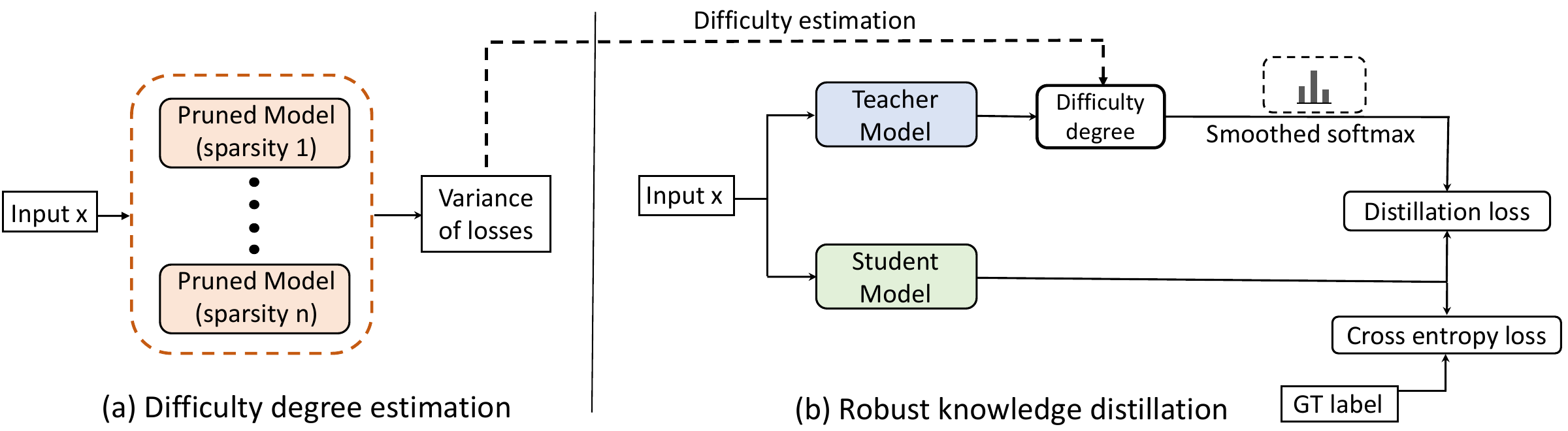}
  \vspace{-4pt}
  \caption{RMC framework for bias mitigation with two-stage training. In the first stage, we feed the training samples to pruned models at different levels of sparsity (ranging from $[0.2 - 0.85]$) as introduced in Section~\ref{sec:Variance-based-Shortcut-Metric}); compute corresponding losses and their variance to estimate the difficulty degree of each training sample. In the second stage, we use the difficulty degree to regularize the teacher network for robust model compression.}
  \label{fig:overall}
  \vspace{-3pt}
\end{figure*}

\vspace{2pt}
\noindent\textbf{Observation 2}: Compressed models tend to assign overconfident predictions to easy samples. One of the potential reasons is that compressed models are more prone to capture spurious correlations between shortcut features in training samples with certain class labels for their predictions~\cite{geirhos2020shortcut,du2021towards}.

\subsection{Variance-based Difficulty Estimation}\label{sec:Variance-based-Shortcut-Metric}
Based on the above observations, we propose a variance-based metric to quantify the difficulty degree of each sample. For each sample in the development set, we calculate its loss at five different levels of pruning sparsity as shown in Figure~\ref{fig:attribution}. We further calculate the variance of the above losses for each sample and rank them based on the variance. Finally, we assign the samples with low variance to the ``easy" subset and rest to the ``hard" one. Comparing our variance-based proxy annotation with the ground truth annotation in~\cite{gururangan2018annotation} gives an accuracy of $82.8\%$. This indicates that the variance-based estimation leveraging pruning sparsity is a good indicator of sample difficulty. This motivates our design of the mitigation technique introduced in the next section.

\section{Mitigation Framework}
In this section, we propose a general bias mitigation framework (see Figure~\ref{fig:overall}), termed as RMC (\underline{R}obust \underline{M}odel \underline{C}ompression), to improve the robustness of compressed models on downstream tasks. Our RMC framework follows the philosophy of task-specific knowledge distillation~\cite{sanh2020movement,jiao2019tinybert}, but with explicit regularization of the teacher network leveraging sample uncertainty. This prevents the compressed model from overfitting in the easy samples that contain shortcut features and helps improve its robustness.  
This regularized training is implemented in two stages.

\subsection{Quantifying Sample Difficulty}
In the first stage, our objective is to quantify the difficulty degree of each training sample.

\vspace{2pt}\noindent\textbf{Variance Computation}:
Following the observations obtained in Section~\ref{sec:Variance-based-Shortcut-Metric}, we first use iterative magnitude pruning to obtain a series of pruned models from BERT-base with different levels of sparsity and then we use the losses of the pruned models at different levels of sparsity to compute their variance $v_i$ for each training sample $x_i$: $v_i=\frac{\sum_{t=1}^n({l}_{i,t}-\bar{l_i})^2}{n}$. We choose five sparsity levels, i.e., $n=5$, that are diverse enough to reflect the difficulty degree of each training sample. Here, samples with high variance correspond to hard ones.

\vspace{2pt}\noindent\textbf{Difficulty Degree Estimation}:
Based on the variance $v_i$ for each training sample $x_i$, we can estimate its difficulty degree as:
\begin{equation}
\small
d_i=\alpha+\frac{1- \alpha}{V_{\max }-V_{\min }} \cdot\left(v_i-V_{\min }\right),
\label{equ:bias-degree}
\end{equation}
where $V_{\min }$ and $V_{\max }$ denote the minimum and maximum values of the variances, respectively. Equation~\ref{equ:bias-degree} is used to normalize the variance of the training samples in the range of [$\alpha$, 1], where $d_i=1$ 
denotes the most difficult training sample, according to our criteria of loss variance. 
Samples with $d_i$ closer to $\alpha$ are treated as shortcut/biased samples. 
Prior work~\cite{niven2019probing} show that the bias behavior of the downstream training set can be attributed to data collection and annotation biases. Since the bias level is different for each dataset, we assign a different $\alpha$ in Equation~\ref{equ:bias-degree} to each training set to reflect its bias level.

\subsection{Robust Knowledge Distillation}

In the second stage, we fine-tune BERT-base on the downstream tasks to obtain the softmax probability for each training sample. We then use the difficulty degree of the training samples (discussed in the previous section) to smooth the teacher predictions. The instance-level smoothed softmax probability is used to guide the training of compressed models through regularized knowledge distillation.

\begin{table*}
\centering
\setlength\tabcolsep{3.5pt}
\scalebox{0.8}{
\begin{tabular}{l c c c c c c c c c c c c |c}
\toprule 
& & \multicolumn{3}{c}{\textbf{MNLI}} & \multicolumn{4}{c}{\textbf{FEVER}} & \multicolumn{4}{c|}{\textbf{QQP}} \\
\cmidrule(l){3-5} \cmidrule(l){6-9} \cmidrule(l){10-13}
\textbf{Sparsity} &\#Param & DEV & HANS &$\mathcal{F}_{bias}$  & DEV & Sym1 & Sym2  &$\mathcal{F}_{bias}$ & DEV & $\text{paws}_{wiki}$ & $\text{paws}_{qqp}$  &$\mathcal{F}_{bias}$ & Average $\mathcal{F}_{bias}$\\
\midrule 
BERT-base       &110M & 84.2 & 59.8 &-  & 86.2 & 58.9 & 64.5&- & 90.9 & 48.9 & 34.7&-&-\\ 
\midrule 
40\% -- Vanilla &65.4M & 84.0  & 54.7& 1.204  & \textbf{86.4} & 57.2 & 64.0& 1.051 & 90.5 & 46.6 & 32.4& 1.049 &1.101\\
-- Distil       &65.4M & 84.1 & 56.2& 1.145   & 86.3 & 58.4 & 64.5& 1.013 & 90.5 & 47.3 & 33.2& 1.032  &1.063\\
-- Smooth       &65.4M & \textbf{84.2} & 56.5& 1.135   & 86.2 & 60.7 & 65.8& 0.937 & \textbf{90.7} & 47.2 & 33.8& 1.036  &1.036\\
-- Focal        &65.4M & 84.0 & 56.7 & 1.122  & \textbf{86.4} & 59.4 & 65.2& 0.981 & \textbf{90.7} & 46.2 & 32.1& 1.060 &1.054\\
-- JTT          &65.4M & 83.8 & 56.3 & 1.132  & 86.2 & 58.1 & 64.9& 1.008 & 90.4 & 47.3 & 33.7& 1.030 & 1.057\\
-- RMC  &65.4M & \textbf{84.2} & \textbf{58.6} & \textbf{1.049}  & 86.1 & \textbf{61.9} & \textbf{66.4}& \textbf{0.897} & 90.4 & \textbf{47.6} & \textbf{34.3}& \textbf{1.023} & \textbf{0.990}\\
\bottomrule 
\end{tabular}}
\vspace{-4pt}
\caption{Generalization accuracy comparison (in percent) and the corresponding $\mathcal{F}_{bias}$ values for iterative magnitude pruning at 40\% sparsity with different \emph{mitigation methods}. The last column indicates average $\mathcal{F}_{bias}$ over three tasks. 
}
\label{tab:iterative-magnitude-pruning-mitigation}
\end{table*}

\vspace{2pt}\noindent\textbf{Smoothing Teacher Predictions}: 
We smooth the softmax probability $\hat{y}_{i}^T$ from the teacher network, according to the difficulty degree $d_i$ of each training sample $x_i$. The smoothed probability is given as: 
\begin{equation}
\small
s_{i,j} = \frac{ (\hat{y}_{i}^T)_j^{d_i}}{\sum_{k=1}^K  (\hat{y}_{i}^T)_k^{d_i}},
\label{equ:smoothing}
\end{equation}
where $K$ denotes the total number of class labels. We perform instance-level smoothing for each training sample $x_i$. If the difficulty degree of a training sample $d_i=1$, then the softmax probability $s_{i}$ for the corresponding sample from the teacher is unchanged. In contrast, at the other extreme as $d_i \rightarrow \alpha$, we increase the regularization to encourage the compressed model to assign less over-confident predictions to the sample. The difficulty degree range is $[\alpha, 1]$ rather than $[0, 1]$ to avoid over-smoothing of the teacher predictions. 

\vspace{2pt}\noindent\textbf{Smoothness-Induced Model Compression}:
We employ the smoothed softmax probability $s_i$ from BERT-base to supervise the training of the compressed models, where the overall loss function is:
\vspace{-3pt}
\begin{equation}\label{eq:loss_self_kd}
\small
\mathcal{L}(x)=(1-\lambda) * \mathcal{L}_1\left(y_i, \hat{y}_{i}^S\right)+\lambda * \mathcal{L}_2\left(s_i, \hat{y}_{i}^S\right),
\end{equation}
where $y_i$ is the ground truth and $\hat{y}_{i}^S$ is the probability of the compressed model. $\mathcal{L}_1$ denotes the cross-entropy loss, and $\mathcal{L}_2$ represents the knowledge distillation loss with KL divergence. Hyperparameter $\lambda$ manages the trade-off between learning from hard label $y_i$ and softened softmax probability $s_i$. Among the different families of compression techniques introduced in Section~\ref{sec:three-families-of-compression}, we directly fine-tune the distilled models using Equation~\ref{eq:loss_self_kd}. For iterative magnitude pruning, we use Equation~\ref{eq:loss_self_kd} to guide the pruning during the fine-tuning process.

\section{Mitigation Performance Evaluation}
In this section, we conduct experiments to evaluate the robustness of our RMC mitigation framework.

\subsection{Experimental Setup}
For all experiments, we follow the same setting as in Section~\ref{sec:Evaluation-Analysis-robustness}, and the same evaluation datasets as in Section~\ref{Evaluation-Datasets}. We use the OOD test set exclusively for evaluation. We compute the variance of samples (outlined in Section~\ref{sec:Variance-based-Shortcut-Metric}) in the in-distribution development set to split it into a shortcut and hard subset. The relative robustness between the hard and easy subset is used to tune the hyperparameter $\alpha$ in Equation~\ref{equ:bias-degree}, where we set $\alpha$ as $0.5$, $0.3$, $0.2$ for MNLI, FEVER, and QQP, respectively. The weight $\lambda$ in Equation~\ref{eq:loss_self_kd} is fixed as $0.9$ for all experiments.

\begin{table*}
\centering
\setlength\tabcolsep{3.5pt}
\scalebox{0.76}{
\begin{tabular}{l c c c c c c c c c c c c |c}
\toprule 
& & \multicolumn{3}{c}{\textbf{MNLI}} & \multicolumn{4}{c}{\textbf{FEVER}} & \multicolumn{4}{c|}{\textbf{QQP}} \\
\cmidrule(l){3-5} \cmidrule(l){6-9} \cmidrule(l){10-13}
\textbf{Sparsity} &\#Param & DEV & HANS &$\mathcal{F}_{bias}$  & DEV & Sym1 & Sym2  &$\mathcal{F}_{bias}$ & DEV & $\text{paws}_{wiki}$ & $\text{paws}_{qqp}$  &$\mathcal{F}_{bias}$ & Average $\mathcal{F}_{bias}$\\
\midrule 
BERT-base  &110M & 84.2 & 59.8  &-      & 86.2 & 58.9 & 64.5&- & 90.9 & 48.9 & 34.7&-&-\\ 
\midrule 
MiniLM -- Vanilla  &66M & 83.1  & 51.4 & 1.309 & 84.2 & 53.4 & 60.7& 1.137 & 89.9 & 46.8 & 31.0& 1.039 &1.162\\
-- Distil &66M & 83.1 & 53.7 & 1.221  & 83.8 & 56.5 & 61.0& 1.052 & 89.6 & 46.7 & 31.8& 1.037 &1.103\\
-- Smooth &66M & 82.7 & 53.8 & 1.206  & 83.7 & 56.9 & 62.1& \textbf{1.017} & 89.4 & 46.8 & 32.2& \textbf{1.032} &1.085\\
-- Focal &66M & 83.2 & 55.6 & 1.145  & 83.8 & 54.7 & 61.4 & 1.081 & 90.3 & 46.8 & 33.2& 1.041 &1.089\\
-- JTT &66M & 82.8 & 55.7 & 1.129  & 83.5 & 53.8 & 61.7& 1.085 & 90.1 & \textbf{47.0} & 32.9& 1.034 &1.083\\
-- RMC &66M & \textbf{83.7} & \textbf{57.8} & \textbf{1.068}  & \textbf{85.3} & \textbf{58.0} & \textbf{63.3}& \textbf{1.017} & \textbf{90.5} & \textbf{47.0} & \textbf{33.4}& 1.038 &\textbf{1.041}\\
\bottomrule 
\end{tabular}}
\vspace{-4pt}
\caption{Generalization accuracy and the $\mathcal{F}_{bias}$ values comparison of different training strategies with and without mitigation on in-distribution development set and OOD test set using MiniLM as the compressed encoder.  
}
\label{tab:knowledge-distillation-mitigation}
\end{table*}

\subsection{Baseline Methods}

We consider the following five baselines. Please refer to Sec.~\ref{More-on-Comparing-Baselines} in Appendix for more details.

\begin{itemize}[leftmargin=*]\setlength\itemsep{-0.3em}
\item \textbf{Vanilla}: This only fine-tunes the base encoder without any regularization.
\item \textbf{Distil} (Task-Specific Knowledge Distillation) \cite{sanh2020movement}: This first fine-tunes BERT-base on the downstream NLU tasks. The softmax probability from the fine-tuned BERT-base is used as the supervision signal for distillation. 

\item \textbf{Smooth} (Global Smoothing)~\cite{muller2019does}: This performs global smoothing for {\em all} training samples with task-specific knowledge distillation, where we use the same level of regularization as in RMC ($d_i= 0.9$ in Equation~\ref{equ:smoothing}). In contrast, RMC uses instance-level smoothing.

\item \textbf{Focal} (Focal Loss)~\cite{lin2017focal}: Compared to cross-entropy loss, focal loss has an additional regularizer to reduce the weight for easy samples and assign a higher weight to hard samples bearing less-confident predictions.

\item \textbf{JTT} (Just Train Twice)~\cite{liu2021just}: This is a re-weighting method, which first trains the BERT-base model using standard cross-entropy loss for several epochs, and then trains the compressed model while up-weighting the training examples that are misclassified by the first model, i.e., hard samples.
\end{itemize}

\subsection{Mitigation Performance Analysis}
We compare our RMC framework with the above baselines and have the following key observations. 

\begin{figure}
  \centering
  \includegraphics[width=1.0\linewidth]{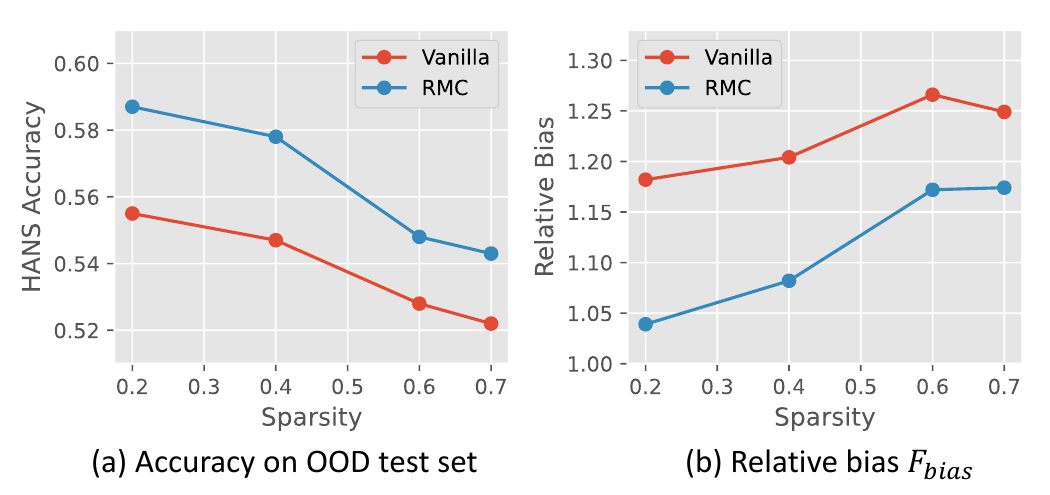}
  \vspace{-7mm}
  \caption{RMC mitigation performance for iterative magnitude pruning at different levels of pruning sparsity for MNLI task. 
  }
  \label{fig:ours-different-sparsity-mnli}
\end{figure}

\vspace{3pt}
\noindent\textbf{Iterative Magnitude Pruning}:
Table \ref{tab:iterative-magnitude-pruning-mitigation} shows the mitigation results of accuracy and relative bias $\mathcal{F}_{bias}$. All mitigation methods are performed with pruned models at $40\%$ sparsity. We observe that task-specific knowledge distillation only slightly improves accuracy on the OOD test set compared to Vanilla tuning, since the teacher model itself is not robust for downstream tasks~\cite{niven2019probing}. Global smoothing further improves generalization accuracy compared to prior methods. Our RMC framework obtains the best accuracy on OOD test set across all the tasks on aggregate. RMC further reduces the average relative bias $\mathcal{F}_{bias}$ by $10\%$ over Vanilla tuning, as shown in Table~\ref{tab:iterative-magnitude-pruning-mitigation}, indicating the benefits of uncertainty-based sample-wise smoothing in terms of improving model robustness. 
For the MNLI task, we also illustrate the mitigation performance of our RMC framework for different levels of sparsity in Figure~\ref{fig:ours-different-sparsity-mnli}. We observe that RMC consistently improves accuracy on OOD HANS while reducing the relative bias $\mathcal{F}_{bias}$ for all levels of sparsity over the Vanilla method.

\vspace{2pt}
\noindent\textbf{Knowledge Distillation}: Table \ref{tab:knowledge-distillation-mitigation} shows the mitigation results of accuracy and relative bias $\mathcal{F}_{bias}$. We observe that RMC significantly improves over MiniLM for OOD generalization leveraging smoothed predictions from BERT-base teacher. With instance-level smoothing in RMC, the generalization accuracy for the compressed model on the OOD test set is significantly closer to BERT-base teacher compared to the other methods. We also decrease the relative bias $\mathcal{F}_{bias}$ in Table~\ref{tab:knowledge-distillation-mitigation} by $10.4\%$ over Vanilla tuning. On the QQP task, RMC simultaneously improves the performance of compressed model on both the in-distribution development set and the two OOD test sets.

\subsection{Further Analysis on Robust Mitigation}
In this section, we further investigate the reasons for the improved generalization performance with RMC with an analysis on the MNLI task. Table~\ref{tab:imrprovement-source} shows the accuracy performance of RMC for model pruning and distillation on the shortcut/easy and hard samples. We observe RMC to improve the model performance on the under-represented hard samples, where it reduces the generalization gap between the hard and shortcut/easy subset by $10.6\%$ at $0.4$ level of sparsity and by $11.3\%$ for knowledge distillation. This analysis demonstrates that RMC reduces the overfitting of the compressed models on the easy samples and encourages them to learn more from the hard ones, thus improving the generalization on the OOD test sets.

\begin{table}
\centering
\scalebox{0.68}{
\begin{tabular}{l c c c c c}
\toprule
    \textbf{Models} &DEV &HANS & Hard (H) & Easy (E) & Gap (E-H) \\
\midrule
MiniLM--Vanilla  &83.1  &51.4   & 73.2 & 90.9   & 17.7 \\
MiniLM--RMC &83.7 &57.8 & 74.9 & 90.6   & 15.7 \\
\midrule
40\%--Vanilla  &84.0  &54.7   & 74.9 & 91.0   & 16.1 \\
40\%--RMC &84.2 &58.6 & 75.9 & 90.3   & 14.4 \\
\bottomrule
\end{tabular}}
\vspace{-4pt}
\caption{Our RMC framework improves accuracy of the compressed models on the hard samples and reduces overfitting on the shortcut/easy samples, leading to reduced performance gap between the two subsets.}
\label{tab:imrprovement-source}
\end{table}

\section{Conclusions} 
In this work, we conduct a comprehensive study of the robustness challenges in compressing large PLMs when fine-tuning in downstream NLU datasets. Furthermore, we propose a general mitigation framework with instance-level smoothing for robust model compression. Experimental analysis demonstrates our framework to improve the generalization and OOD robustness of compressed models for different compression techniques, while not sacrificing the in-distribution performance.

\section*{Limitations}
First, we study the shortcut learning/bias problem and OOD generalization of model compression techniques, exclusively focusing on the two most widely used families of compression techniques, including knowledge distillation and pruning. Our empirical analysis indicates that these two families of compression techniques suffer from the low generalization issue. However, other types of compression technique, such as matrix decomposition and quantization, are not discussed in this work. Studying the whole compression techniques is a challenging topic and will be investigated in our future research. Second, our RMC framework needs to calculate the variance of losses for each training sample, thus requiring additional training time. Training efficiency can be further improved by implementing parallel training or more efficient ways of calculating sample difficulty, which will also be studied in our future research.

\bibliography{acl2020}
\bibliographystyle{acl_natbib}

\clearpage
\appendix

\section{More Details of Pruning Methods}\label{Details-of-Pruning-Methods}

In this section, we introduce more details about the compression techniques studied.

\noindent\textbf{knowledge Distillation}: For a fair comparison, we do not compare with TinyBERT~\cite{jiao2019tinybert} and MobileBERT~\cite{sun2020mobilebert}, since TinyBERT is fine-tuned with data augmentation on NLU tasks, and MobileBERT is distilled from BERT-large rather than BERT-base.

\noindent\textbf{Magnitude Pruning}: It is based on the overparameterization assumption of pre-trained language models~\cite{xu2021rethinking,huang2021sparse}.
For iterative magnitude pruning, we freeze all the embedding modules and only prune the parameters in the encoder (i.e., 12 layers of Transformer blocks). After pruning, the pruned weight values are set to 0 to reduce the amount of information to store. Unlike the LTH version, we consider standard magnitude pruning without using rewinding.

\vspace{3pt}
\noindent\textbf{Structured Pruning}:
To calculate the importance, we follow~\cite{michel2019sixteen,prasanna2020bert} and calculate the expected sensitivity of the attention heads to the mask variable $\xi^{(h, l)}$:
$
I_{h}^{(h, l)}=E_{x \sim X}\left|\frac{\partial \mathcal{L}(x)}{\partial \xi^{(h, l)}}\right|,
$
where $I_{h}^{(h, l)}$ denotes the contribution score of the attention head $h$ in layer $l$, $\mathcal{L}(x)$ represents the loss value for the sample $x$, and $\xi^{(h, l)}$ is the mask of the attention head $h$ in layer $l$. After obtaining the contribution scores, the attention heads with lowest score $I_{h}^{(h, l)}$ are pruned. 

\section{More on Evaluation Datasets}\label{More-on-Evaluation-Datasets}

In this section, we introduce more details about the three benchmark datasets.

\noindent\textbf{MNLI}: This task aims to predict whether the relationship between the premise and the hypothesis is contradiction, entailment, or neutral. It is divided into a training set and development set with $392,702$ and $9,815$ samples, respectively.

\vspace{3pt}
\noindent\textbf{FEVER}:
The task is to predict whether the claims support, refute, or not-have-enough-information about the evidence. Recent studies indicate that there are strong shortcuts in claims~\cite{utama2020towards}. It is divided into a training set and a development set with $242,911$ and $16,664$ samples, respectively. 

\vspace{3pt}
\noindent\textbf{QQP}:
It is divided into a training set and a development set with $363,846$ and $40,430$ samples, respectively. 

\section{More on Comparing Baselines}\label{More-on-Comparing-Baselines}

In this section, we introduce more details on comparing baselines.

\noindent\textbf{Distil and Smooth}: For both baseline methods, we use a loss function similar to that of Equation~\ref{eq:loss_self_kd}. We fix the weight $\lambda$ to 0.9 for all experiments, to encourage the compressed model to learn more from the probability output of the teacher network. A major difference between the two baselines is that Smooth has an additional smoothing process involved during the fine-tuning process.

\vspace{3pt}
\noindent\textbf{Focal Loss}:\,
The original focal loss function is: $
\mathrm{FL}\left(p_{\mathrm{i}}\right)=-\left(1-p_{\mathrm{i}}\right)^{\gamma} \log \left(p_{\mathrm{i}}\right)
$. 
Our implementation is as follows: $$
\mathrm{FL}\left(p_{\mathrm{i}}\right)=-\frac{\left(1-p_{\mathrm{i}}\right)^{\gamma}}{\frac{1}{N}\sum_{k=1}^N \left(1-p_{\mathrm{k}}\right)^{\gamma} } \log \left(p_{\mathrm{i}}\right).
$$
The hyperparameter $\gamma$ controls the weight difference between hard and easy samples, and is fixed at 2.0 for all tasks. We use the denominator to normalize the weights within a batch, where $N$ is the batch size. This is used to guarantee that the average weight for a batch of training samples is 1.0. As such, the weight for the easy samples would be down-weighted to lower than 1.0, and the weight for hard samples would be up-weighted to values larger than 1.0.

\vspace{3pt}
\noindent\textbf{JTT}:\, This is also a reweighting baseline that encourages the model to learn more from hard samples. The hyperparameter $\lambda_{up}$ in ~\cite{liu2021just} is set to 2.0. We also normalize the weights so that the average weight for each training sample is 1.0.

\section{Running Environment}
For a fair evaluation of the robustness of compressed models, we run all experiments using a server with 4 NVIDIA GeForce 3090 GPUs. All experiments are implemented with the Pytorch version of the Hugging Face Transformer library.

\section{The Capacity Issue}
One natural speculation about the low robustness of compressed models is due to their low capacity (i.e., smaller size). To disentangle the two important factors that influence model performance, i.e., low capacity and compression, we compare distilled models with Uncased-l6, which is trained only using pretraining. The results are given in Table~\ref{tab:knowledge-distillation}. The results indicate that Uncased-l6 has better generalization ability over the MNLI and FEVER two tasks. Take structured pruning as an example; although the three pruned models in Table\ref{tab:structured-pruning} have the same model size, their generalization accuracy is different. These results indicate that the low robustness of compressed models is not entirely due to their low capacity, and compression plays a significant role.

\section{MNLI Easy and Hard Subsets}
The authors train a hypothesis-only model and use it to generate predictions for the whole development set~\cite{gururangan2018annotation}. Samples that are given correct predictions by the hypothesis-only model are regarded as easy samples, and vice versa. The easy subset contains $5488$ samples, and the hard subset contains $4302$ samples.

\end{document}